%% file: main.tex
\documentclass[letterpaper, 10 pt, journal, twoside]{ieeetran}
\ifCLASSINFOpdf
\else
\fi

\usepackage{amsmath}
\usepackage{graphicx}
\usepackage[vlined,ruled]{algorithm2e}
\usepackage{color}
\usepackage{multirow}
\usepackage[table,xcdraw]{xcolor}
\usepackage{float}
\usepackage{caption}
\usepackage{subcaption}
\usepackage{xspace}
\usepackage{balance}
\usepackage{makecell}
\usepackage{tabularx}
\usepackage{enumitem}
\usepackage{hyperref}
\usepackage{fancyhdr}
\usepackage{footmisc}

\newcommand{\system}{DoRO\xspace}
\DeclareMathOperator*{\argmax}{argmax}

\begin{document}

%\title{An efficient and reliable robot packing system}
%\title{When robot fault recovery meets \\cuboid object packing}
%\title{Jampacker: Cuboid object packing meets \\robot fault recovery}
%\title{Jampacker: Towards efficient and reliable \\ robotic bin packing for cuboid objects}
\title{\system: Disambiguation of referred object for embodied agents}


\author{Pradip Pramanick, Chayan Sarkar, Sayan Paul, Ruddra dev Roychoudhury, and Brojeshwar Bhowmick \\
Robotics \& Autonomous Systems, TCS Research, India
}
\maketitle

\input{0_abstract}


% \fi
\IEEEpeerreviewmaketitle

\input{1_intro}

\input{2_related}
\input{3_overview}

\input{4_details}
\input{5_evaluation}
\input{6_conclusions}

\bibliographystyle{IEEEtran}
\bibliography{main}
\end{document}

%% file: 0_abstract.tex
\begin{abstract}
Robotic task instructions often involve a referred object that the robot must locate (ground) within the environment. While task intent understanding is an essential part of natural language understanding, less effort is made to resolve ambiguity that may arise while grounding the task. Existing works use vision-based task grounding and ambiguity detection, suitable for a fixed view and a static robot. However, the problem magnifies for a mobile robot, where the ideal view is not known beforehand. Moreover, a single view may not be sufficient to locate all the object instances in the given area, which leads to inaccurate ambiguity detection. Human intervention is helpful only if the robot can convey the kind of ambiguity it is facing. In this article, we present \system (\emph{\underline{D}isambiguation \underline{o}f \underline{R}eferred \underline{O}bject}), a system that can help an embodied agent to disambiguate the referred object by raising a suitable query whenever required. Given an area where the intended object is, \system finds all the instances of the object by aggregating observations from multiple views while exploring \& scanning the area. It then raises a suitable query using the information from the grounded object instances. Experiments conducted with the AI2Thor simulator show that \system not only detects the ambiguity more accurately but also raises verbose queries with more accurate information from the visual-language grounding.

\end{abstract}

%% file: 1_intro.tex
\section{INTRODUCTION}
There is an upsurge of robots that are used as assistants/companions rather than just high-precision tools~\cite{cooper2020ari, pramanick2018defatigue}. The expectation is that they can act autonomously (to some extent if not fully) so that any novice user can also use them. Natural human-robot interaction certainly plays an important role in that~\cite{matuszek2013learning, williams2015going, pramanick2020decomplex}.

\textbf{Motivation.}
Imagine a daily life scenario where we ask a robot to help us with some task. E.g., ``can you bring me the red cup from the dining table'' or ``please place the flower in the vase near the couch''. In the first case, first, the robot needs to identify the \textit{red cup on the dining table}, and in the latter, it is the \textit{vase placed beside the couch}. However, while searching for the red cup, one may find only a black cup on the dining table or a red cup at the nearby counter, or none at all. In other words, there is no unique object that matches the description. In such a case, humans tend to ask a suitable question to clarify the scenario and the next course of action. In this work, we define this as a new task for a cohabitant embodied agent and develop a system that would enable an agent to perform such tasks.

\begin{figure}[t!]
    \centering
    \includegraphics[width=\linewidth]{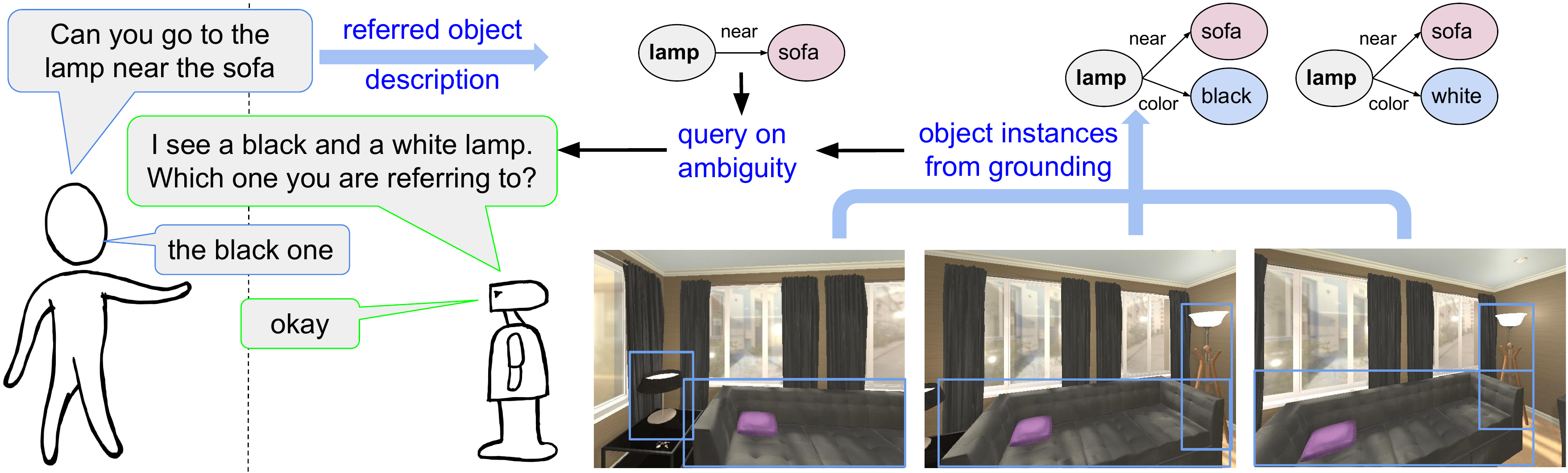}
    \caption{An illustration of how disambiguation works in \system.}
    \label{fig:representative}
% \vspace{-0.6cm}
\end{figure}

\textbf{Problem description.}
%The goal of natural language understanding (NLU) in robotics applications is primarily aimed toward understanding the task intention, regardless of how such intentions are expressed. Additionally, this requires grounding the task in the environment the robot is operating. Most existing works on task understanding \cite{brawer2018situated, pramanick2020decomplex} assume the availability of a knowledge base (KB) of the environment that is sufficient to ground the task; hence limited to only intention understanding. However, a KB can be utilized for location reference and it would not be sufficient if there are multiple instances of the object available in the same region. 
Using an object detection algorithm, the referred object instance(s) can be identified, but the ambiguity arising from multiple matching or mismatching instances can not be resolved. Existing work on ambiguity detection \cite{zhang2021invigorate, pramanick2021talk} is limited to processing a single image primarily for a non-mobile robot in a table-top scenario. However, an embodied agent (often being mobile) needs to process multiple views of the environment to ground the intended object. It is almost impractical to predetermine the best view to ground the object and determine if there is any ambiguity. Thus, multiple views of the environment must be processed. In case of ambiguity, human intervention is required and a query can be raised for the same. However, unless the generated query conveys the ambiguity in a meaningful way, the human being would not be able to help resolve it. 

\textbf{Our Approach.}
In this work, we introduce a new task for an embodied agent that aims to identify ambiguity in grounding a referred object through exploration and raise a descriptive query that can help to disambiguate the target object. Our system is applicable where a human being may or may not share the same physical environment, but is aware of the environment and its constitutes in general. Moreover, the human may or may not have visual feedback of what the robot is currently viewing. Thus, any instruction to the robot is presumptive. We develop a novel system called \system (Fig.~\ref{fig:representative}) to solve the task. In order to identify the referred object that matches the description provided in the natural language instruction, \system creates an input graph from the instruction and aggregates observations from multiple views of the environment to find the unique instance graph(s) of the referred object. The root node of a graph represents the referred object and the edges point to self attributes (such as color, material, etc.) and relational attributes of the object (with respect to other objects in its vicinity). Finally, \system uses a graph discrimination algorithm to find ambiguity and generates query if there is ambiguity.
%\system needs to handle multi-modal data. Thus, we form a input graph of the referred object from the natural language instruction by training a transformer based network. We also create instance graph(s) of the detected objects using our efficient scene aggregation algorithm that takes input from object detector. Then, we use our graph discrimination algorithm to compare the input and instance graphs and check if there is a single instance of the target object that matches the description exactly; any other situation is broadly termed as ambiguity. Finally, we raise a query for the human being by describing the scenario in case there is ambiguity. 

\noindent Our contributions can be summarized in the following.

\begin{itemize}[leftmargin=*]
\item Conceptually, we formulate the object disambiguation problem for an embodied agent that requires natural language understanding, visual scene understanding, ambiguity detection, and query generation. To the best of our knowledge, this is the first attempt that detects ambiguity in object instance grounding from multiple views of the environment and generates a context-specific query for disambiguation.

\item Technically, we propose a novel system that is well-orchestrated using deep learning-based and deterministic algorithms-based sub-systems. We learn to form object graphs from any natural language phrase by training a BERT-based phrase-to-graph network. We develop a multi-view aggregation algorithm to merge the instance graphs across multiple frames to find the unique object instances. We develop a graph discrimination based deterministic algorithm that generates accurate queries on ambiguity.

\item Empirically, we conduct experiments on AI2Thor, an interactive embodied AI simulator to show the efficacy of our system. Compared to a baseline system, \system achieves 2X more accuracy in describing the ambiguous scenario.

\end{itemize}

%% file: 2_related.tex
\section{RELATED WORK}
One of the fundamental tasks for an embodied agent is to \textbf{i}dentify and \textbf{l}ocate \textbf{o}bjects (ILO) within the environment in which it is residing. There are many applications that require ILO as a capability, e.g., visual question answering~\cite{antol2015vqa, shih2016look}, visual semantic navigation~\cite{li2019visual, chaplot2020object}, interactive object picking~\cite{zhang2021invigorate, hatori2018interactively, novkovic2020object}, etc.

Natural language understanding and grounding is a key feature of any embodied agent. Prabhudesai \textit{et al.}~\cite{Prabhudesai_2020_CVPR} proposed a language grounding method for locating objects in 3D without any specific application. Pramanick \textit{et al.}~\cite{pramanick2019enabling} proposed a system that focuses on task instruction understanding and grounding in the environment. Similarly, Dongcai \textit{et al.}~\cite{dongcai17integrating} also proposed a method that mostly focuses on instruction understanding for plan generation. However, natural language is ambiguous and dialogue is used to disambiguate a task instruction and ground the user intention~\cite{wu2020tod, pramanick2019your}. The methods vary from slot filling approach~\cite{thomason2019improving} to a reasoning approaching utilizing the existing knowledge~\cite{chen2020enabling}. However, even after understanding the intended task, the robot may face ambiguity in ILO. 

In practice, a robot may face difficulty in grounding the referred object during execution. To tackle this, a visual understanding of the environment against the linguistic input is followed by many researchers. Though several works have specifically focused on the visual grounding of natural object descriptions~\cite{cohen2019grounding,magassouba2019understanding,sadhu2019zero}, they do not tackle ambiguity. Moreover, predominant approaches of end-to-end training for visual grounding pose challenge to integrate a dialogue system, which generally require fine-grained scene analysis. Although visual question-answering systems can perform fine-grained scene analysis~\cite{johnson2017clevr, teney2018tips}, they are limited to answering questions, as opposed to generating a specific query describing the cause of the ambiguity. Recent works on ambiguity resolution for picking task~\cite{zhang2021invigorate,shridhar18,yang2022interactive} detect ambiguity in object grounding and raise a query if detected. However, these methods are mostly limited to table-top setup, where the agent assesses the scene from a single view. In the case of a mobile agent, such a (best) view is not known beforehand. As a result, aggregation of observations from multiple views is essential in order to determine the ambiguity correctly. 

Although some recent works \cite{chen2020scanrefer, yuan2021instancerefer} have tried to perform 3D visual grounding of natural language descriptions directly on aggregated point-cloud, they do not deal with ambiguous scenarios. They use detailed, specific descriptions to localize objects and have no dialogue system to assist the robot when it fails to ground an object. Moreover, their training is end-to-end, which requires large annotated 3D datasets and is also a sample-inefficient approach. Also, most of these systems are not real-time and consume a lot of memory and computing resources to do panoptic segmentation on entire scene point clouds, followed by post-processing methods like clustering, fusion, etc. This hinders their direct usage in our front-end instance graph generation module. In contrast, we follow a 2-stage process where any 2D object detector can be used in the multi-view aggregation module, which only aggregates  relevant information and generates instance graphs for ambiguity resolution. This enables us to make our system lightweight, real-time, and also helps in generalization to diverse scenarios.

\begin{figure*}
    \centering
    \includegraphics[width=\linewidth]{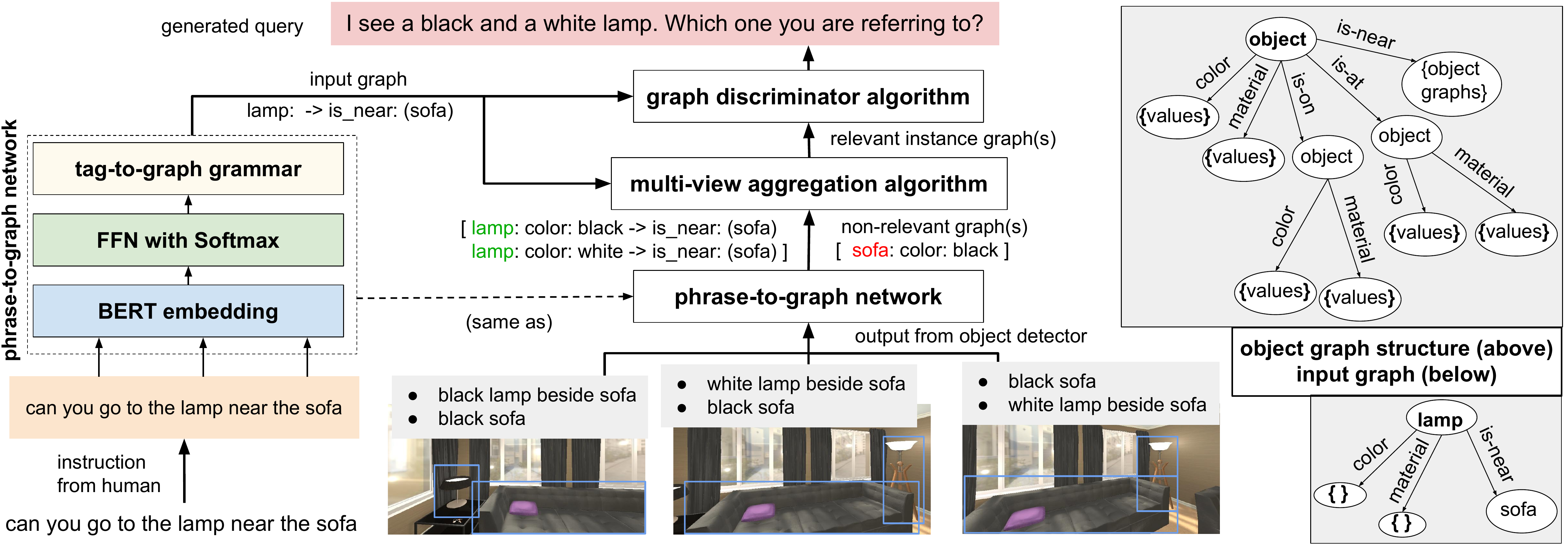}
    \caption{Overall system architecture of \system, which constitutes three major components -- phrase-to-graph network, multi-view aggregation algorithm, and graph discriminator algorithm along with the structure of our object graph.}
    \label{fig:doro_model}
%\vspace{-0.6cm}
\end{figure*}

%% file: 3_overview.tex
\section{\system in a nutshell}

% \begin{figure}[t!]
%     \centering
%     \includegraphics[width=\linewidth]{figs/system_overview_doro.pdf}
%     \caption{The major building blocks of \system and how it can be integrated with other sub-systems for deployment.}
%     \label{fig:overview}
% \end{figure}

In this section, we provide an overview of \system before describing the finer details of the system. Fig.~\ref{fig:doro_model} shows three building blocks of \system -- phrase-to-graph network, multi-view aggregation algorithm, and graph discriminator algorithm. \system takes task instruction from a human instructor as natural language text. However, it can be extended for audio-based interaction by converting the audio input to text using any state-of-the-art automatic speech recognition (ASR) system. We assume the instruction contains a task and a referred object with or without attributes. Firstly, \system identifies the referred object and its attributes by parsing the instruction into a graph, i.e., the input graph using our phrase-to-graph network. Given the input graph, thereafter, \system attempts to identify all instances of the referred object in the area. The robot does not have any prior knowledge about where exactly the object(s) is(are) located inside the room or how many instances of the object are there if at all. Moreover, if the area is large enough, a single view from the robot's camera may not be sufficient to locate all the objects. Thus the robot performs an exploration of the area to collect observations. Each observation records the robot's relative pose change and the RGB-D image. The RGB images corresponding to each view are processed to identify the objects and their attributes within a set of 2D bounding boxes. We do so by generating a natural language description (caption) of a given bounding box. In this article, we use existing object detection methods. From the generated caption of a single observation, we form the instance graph by using our phrase-to-graph network. Given this history of observations, \system projects the masked point clouds of the 2D bounding boxes into a grid-map and approximately localizes the possible instances of the referred object. This process helps to merge the instance graphs across observations and provides only the unique instance graph(s). Finally, given the input graph and the unique instance graph(s), \system decides if an unambiguous grounding of the referring expression of the object is possible by using our graph discrimination algorithm. In case of ambiguity, the algorithm also generates a query by using the mutually discriminative attributes of the instance graphs.

%% file: 4_details.tex
\section{\system in details}

\subsection{Phrase-to-graph network}
\label{sec:parsing}
Throughout the pipeline of \system, we maintain a semantic representation of an object in the form of a graph. Thus we convert any natural language description of an object into a graph. In the following, we formally define the general structure of an object graph. 

An object graph $g$ is a tree with the class of the referred object at the root, $r(g)$. The root node has children that encode the attribute types of the referred object. There are two types of attribute nodes -- self attributes $at_S$ and relational attributes $at_R$. Each $at_S$ has a single child as a leaf node $av_S$ denoting the value (token) of the self attribute type. Each $at_S$ encodes a physical (self) attribute type of the object such as color, material, etc. The corresponding leaf node holds the token of the type, e.g., red, black, white, etc., for color, and wooden, metal, glass, etc., for the material. Each $at_R$ describes a spatial relationship with another object node. An $at_R$ has a child node denoting the class of the object and its own $at_S$ and $at_R$ nodes.

We take a two-stage pipeline to convert a natural language text into an object graph. In the first stage, a sequence tagger jointly predicts the referred object class, the attribute types, and their values. Given a sequence of tokens $\{t\}_{i=0}^n$, the sequence tagger predicts a sequence of labels $\{l\}_{i=0}^n$ from the set of symbols $C$,
\vspace{-0.15cm}
\[C=\Big \{r(g), \{at_S\} \cup \{at_R\}, av_S, av_R, o \Big \} , \] where $o$ denotes a non-relevant token and $ \{at_S\} \cup \{at_R\}$ is the union of all self and relational attribute types considered. The model supports arbitrary self and relational attributes by building the set $ \{at_S\} \cup \{at_R\}$ from the training data. To handle nodes with multi-token spans, we use the well-known BIO tagging scheme, expanding $C$ with \textit{B-} and \textit{I-} prefixes. We model the sequence tagger as a transformer network. We obtain the contextual hidden representation $h_i$ for each input token $t_i$ using a transformer-based pre-trained BERT model. The hidden vector is fed to a feed-forward layer with softmax to produce a probability distribution over $C$ for each token. Thus, we obtain the label sequence $l_{1:n}$ as,
\[ h_i = [ BERT(t_i)]\]
\vspace{-0.4cm}
\[  l_{1:n} =\argmax_{l_i \in C} P(l_i | FFN(h_i)) .\]
Given the predicted label sequence, we construct the object graph using a deterministic top-down parsing algorithm. Following the formal structure of an object graph defined above, we formulate a parsing grammar (tag-to-graph grammar in Fig.~\ref{fig:doro_model}. A simplified form of the grammar is as the following,
\begin{align*} 
r(g) \rightarrow at_{S}^* | at_{R}^*,    & at_S \rightarrow av_S, \\
at_R \rightarrow av_R ,  & av_R \rightarrow at_{S}^* | at_{R}^* .
\end{align*}
The sequence tagger detects the primary refereed object $r(g)$ in the text. It also performs a many-to-one mapping from tokens to nodes in the graph, as there can be multiple object classes mentioned in the text and multiple instances of the same word that denotes attributes of different objects, e.g., ``a white lamp near a white table". The parser builds the tree, starting with the token labeled as $r(g)$ in a top-down manner, by retrieving the labels from a stack and creating the appropriate nodes and edges according to the grammar.

\begin{figure*}
    \centering
    \includegraphics[width=\textwidth]{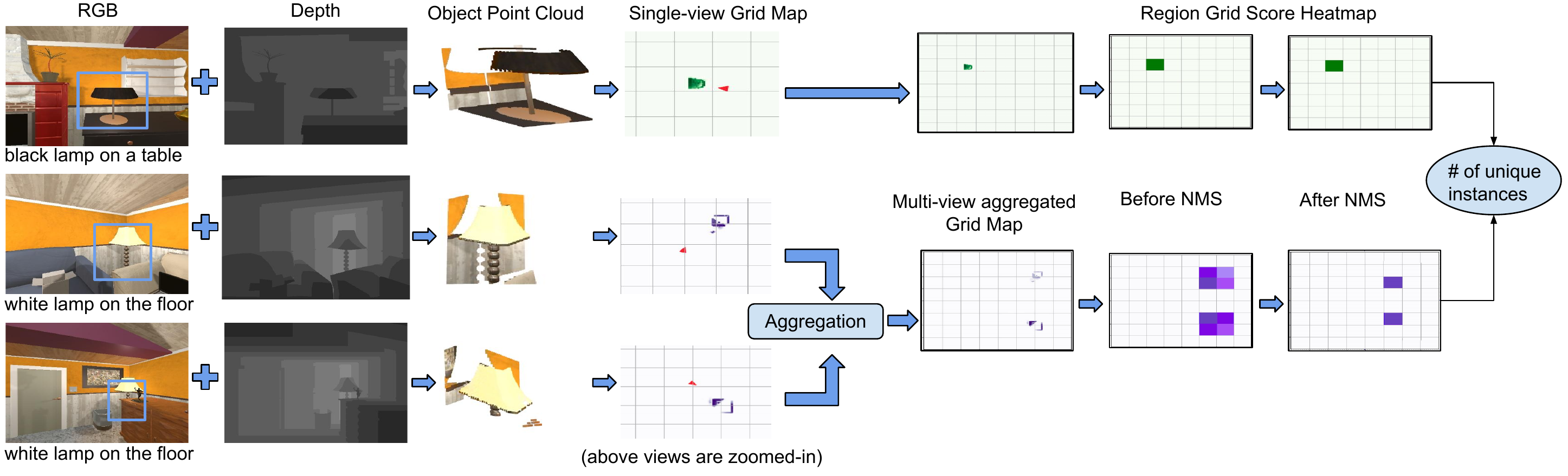}
    \caption{Illustration of how our multi-view aggregation algorithm works.}
    \label{fig:projection-viz}
%\vspace{-0.6cm}
\end{figure*}

\subsection{Multi-view aggregation algorithm}
\label{sec:target_object_detection}
Given the sequence of RGB-D images and absolute poses of the agent obtained from local exploration, our objective is to uniquely identify a particular object instance along with its spatial and physical attributes. For each RGB-D frame $f_t$ and an absolute camera pose at time-step $t$, each point $p_i(u_i,v_i)$ on the image frame $f_t$ has an associated depth value $d_i$. We use an off-the-shelf 2D object detector to detect relevant objects in the RGB frame $f_t$. Having knowledge of the camera intrinsic matrix $K$, we can re-project each image point $p_i$ lying inside the bounding box of the detected object into the camera coordinates 3D space as $P^c_i$. \\

$[x^c_i\;\;\;\;  y^c_i\;\;\;\; z^c_i]^T=$
$K^{-1}$
$[u_i\;\;\;\; v_i\;\;\;\; d_i]^T$
\\

Then from the camera coordinates space, the point cloud $P^c$ is transformed to the world coordinates space $P^w$ using the absolute pose of the agent at that time-step. \\

$[x^w_i\;\;\;\; y^w_i\;\;\;\; z^w_i]^T$
$=$
$ \begin{bmatrix}R & t \\ 0 & 1\end{bmatrix} $
$[x^c_i\;\;\;\; y^c_i\;\;\;\; z^c_i]^T$
\\
where $R$ and $t$ are the rotation matrix and translation vectors corresponding to the absolute pose with respect to the world origin. The bounding box of a detected object also contains the background of an object. Even when a segmentation mask is available, it can be inaccurate near boundaries. So we use a soft-mask approach where the points inside the bounding box of the object are weighted using a 2D Gaussian function to weigh the centers of the objects more than the boundaries and the background.
\[w_{u,v} = {1\over{2*\sigma_u\sigma_v}}
\exp{{-1\over2}\left[{(u-u_c)^2\over\sigma_u^2}
+{(v-v_c)^2\over\sigma_v^2}\right]}, \]
where $(u_c,v_c)$ is the center of the bounding box in image coordinates. The values of $\sigma_u$ and $\sigma_v$ are chosen empirically. We find that it helps suppress erroneous predictions as the edges of a bounding box are less likely to lie on the object. As each 2D point $p_i$ has a corresponding 3D point $P^w_i$ in the world coordinates space, so each $P^w_i$ will also have an associated weight $w_i$. The object point cloud is then discretized into a voxel grid and its top-down bird's eye view (BEV) projection is taken to form a 2D occupancy grid map. The average weight of the points in a grid cell is assigned as the weight of that grid cell.

For each detected object in each successive frame in the RGB-D sequence, the natural language description of the object (if the 2D object detector is a dense caption predictor) is converted to an object graph using the same sequence tagging followed by the parsing pipeline described in Section~\ref{sec:parsing} or the attributes of the object are formally represented as an object graph (if using an object-cum-attribute detector). If the object graph formed is a new one then it is assigned an auto-incremental unique ID $oid_i$ or else if it is similar to an existing object graph in the hashmap (or database), then it is assigned the same ID.

For efficient storage and retrieval, we store the object graph ids, grid map cells, and their weights in 2 hash-map data structures which have an expected storage and retrieval time complexity of O(1). The first Hashmap stores the object ID $oid_i$ as key and the list of grid cell coordinates $(x^g_j,y^g_j)$ of the BEV projection of the object point cloud as values where $0<=j<=m$ and m is the total number of grid cells pertaining to an object, e.g.,  $oid_1 \xrightarrow[]{} [(x^g_1,y^g_1),(x^g_2,y^g_2),(x^g_3,y^g_3), ...]$.

The second Hashmap is a sparse occupancy map data structure where the grid cell coordinates $(x^g_j,y^g_j)$ are the keys and their weights $w^g_i$ and the total number of detections (frequency) $freq^g_i$ of the same object graph till the current time-step are the values, e.g., $(x^g_1,y^g_1) \xrightarrow[]{} (w^g_i, freq^g_i)$.

So for each unique object graph having a unique id, we generate the dense occupancy 2D grid maps from the corresponding set of grid cells. Now, there can be multiple instances of the same object/object graph in the scene, e.g., a white lamp on the floor. To find out the number of unique instances, we formulate a novel approach that analyzes the grid cell weight estimates to find the local maxima that approximately localize object instance graphs $\{Ig_i\}_{i=0}^n$, where $n$ is the number of object instances in the explored area. Firstly, we define a physical region $R=(dx,dy)$ on the grid map with the following assumptions.
\begin{enumerate}[leftmargin=*]
    \item A region $R$ contains at most one instance graph, $Ig_i$. In other words, we assume no ambiguity within a region of size $(dx, dy)$, where $dx$ and $dy$ are the size of $R$ along the x and y-axis on the occupancy grid-map.
    \item Each $Ig_i$ is localized by $m$ regions that are mutually neighbors, where $m\geq1$. Irrespective of object classes, we do not assume an upper bound of $m$.
\end{enumerate}

Next, we obtain an occupancy score for all the regions by summing the grid cell weights within a region and then normalizing,
\[ O(R)_k =  \frac{\sum_{j=1}^{dx \times dy} w^g_i}{\sum_{k=0}^{(d1\times d2)/(dx \times dy)} \sum_{j=1}^{dx \times dy} w^g_i} ,\]
where $d1 \times d2$ is the size of the grid map. The normalized occupancy score distribution over the grid map approximates the probability of finding any instance of a given object class, i.e the root of the object instance graph in a region, $P(r(g) \in R_k)$. However, as we do not have a prior of $m$, we devise an algorithm to merge neighboring regions, while pruning noisy regions from the map. Algorithm~\ref{algo:nmm} shows the merging and noise suppression process.

After this process, we get the number of instances of each unique object graph from each region grid map. So, the total number of unique instances of an object would be the sum of all the instances of all the unique object graph(s) pertaining to an object. But sometimes, due to noisy 2D object/attribute/caption detection, there can be multiple different object graphs for the same object instance. To handle such scenarios, we stack the region grid maps for each unique object graph on top of the other, and then max-pool along the stacked dimension to get the final number of unique instances of a referred object.
% \vspace{-0.3cm}
\begin{algorithm}
\small
\LinesNumbered
\SetAlgoLined \DontPrintSemicolon
\SetKwInOut{Input}{Input}\SetKwInOut{Output}{Output}
\SetKwInput{Initialize}{Initialization}
\Input{Occupancy score matrix as vector :O}
\Initialize{merged=$\emptyset$}
\SetKwProg{nmMalg}{Algorithm}
\nmMalg{\algo}{
Obtain a vector of region indices $\Vec{RI}$ by sorting $O$ in descending order \;
\For{$index \in \Vec{RI}$ and index $\notin$ merged}
    {
    %\tcp{Noise suppression}
    \If{$O(index) < \gamma $}
        {$O(index)=0$}
    \Else{
        %\tcp{Neighbour merging}
        next-best= $\Vec{RI}$[index+1] \;
        \If{is-neighbour(index,next-best)}
            {O[next-best] = O[index] \;
            merged.add(next-best)  \;}
        \Else{
            \For{j in merged}{
                \If{is-neighbour(j,index)}{
                    O[index] = O[j] \;
                    \textbf{break} \;
                }
            }
        }
    }    
}
\Output{Merged region scores}
}
\caption{Greedy non-maximal region merging}
\label{algo:nmm}
\end{algorithm}
%\vspace{-0.6cm}

\subsection{Graph discriminator algorithm}
\label{sec:dialouge}
The graph discriminator algorithm serves both the purposes of ambiguity detection and query generation jointly.
\subsubsection{Ambiguity identification }
Given an input graph $g$ and a set of unique instance graphs obtained from the metric grid-map $\{Ig\}$, we compute the set of discriminative instance graphs $\{Ig'\}$. For each generated instance graph, we compute a pairwise set difference with $g$ and remove empty results.
\vspace{-0.1cm}
\[ \{Ig'_i = (g - Ig_i), Ig'_i \neq \emptyset \}_{i=0}^{n} .\]
By generating the pairwise symmetric difference set, we classify one of the four states shown in Table~\ref{tab:question_templates}. Therefore we decide on an exact match if $\{Ig' = \emptyset\}$, i.e., the \textsc{confirm} state. Otherwise, the cardinality  of the set is used to decide between a mismatch and an ambiguity, i.e., 
\begin{align*}
\small
(|\{Ig'\}| = 1 ) \Rightarrow \textsc{inform-mismatch}, \\
\small
(|\{Ig'\}| > 1 ) \Rightarrow \textsc{inform-ambiguity}.
\end{align*}
In case the set of instance graph is empty, i.e., $\{Ig\} = \emptyset $, we decide on the \textsc{inform-missing} state.

\subsubsection{Query generation}
The question is crafted to convey the robot's partial understanding from the exploration and describe the source of ambiguity/mismatch in natural language. To generate such pin-pointed questions, we resort to a set of question templates. Furthermore, we randomize parts of the templates to generate the questions dynamically and make them seem natural. 
\begin{table}[t]
    \centering
    \caption{Templates for question generation. Slots are shown within the parenthesis [], \textsuperscript{+} denotes the slot can be repeated.}
    \begin{tabular}{|l|p{5.2cm}|}
    \hline
    \textbf{State} & \textbf{Question template} \\ \hline
         \textsc{inform-mismatch} & I found one [\textit{graph description}] \newline [random mismatch-suffix]. \\ \hline
         \textsc{inform-ambiguity} & I found one [\textit{graph description}]\textsuperscript{+} [, and]\textsuperscript{+} \newline [random wh-suffix] \\ \hline
         \textsc{confirm} & [random acknowledgement phrase] \\ \hline
         \textsc{inform-missing} & I could not find that. \\ \hline
    \end{tabular}
    \label{tab:question_templates}
%\vspace{-0.6cm}
\end{table}
Table \ref{tab:question_templates} shows the question templates and their mapping with the states. Each template contains slots that are replaced by a natural language description of an object graph. To generate the description of object graphs, we augment the parsing grammar with rules that annotate the edges of the object graph (tree) with surface forms, $edge \in at_S \cup at_R$. For example, the \textit{is-on} relationship is converted to the surface form \textit{`on top of'} and \textit{color} is converted to an empty string. The nodes already contain surface forms as tokens. Following the English grammar, we maintain an ordering of the edges, such that any $at_R$ edge is always on the right of an $at_S$ edge. Therefore a pre-order traversal of the tree produces the description of the object graph in English.

%% file: 5_evaluation.tex
\section{EVALUATION}
We evaluate \system using the \textit{ai2thor} simulator~\cite{ai2thor}. The simulator allows an embodied agent to explore within a geofence of different rooms and capture RGBD observations. It also provides object metadata and rendering of an oracle object detector. In the following, we describe the construction of the evaluation dataset and experiments performed on it.
%\vspace{-0.15cm}
\subsection{Dataset}
We build the dataset focusing on some important assumptions. We describe the components of the dataset below.
\paragraph{Object types}
As we assume our approach is agnostic to the object shape and size, we select a subset of the ai2thor provided object types accordingly. We select a mixture of small, medium, and large-sized objects such as cup, laptop, and sofa. We also select both convex objects such as book, pillow, etc., and non-convex objects such as chair, plant, etc. %Fig.~\ref{fig:object-room-distribution} shows a diagram of the object types with their occurrence in the dataset.
We purposefully exclude objects that always occur exactly once (e.g., television, fridge) to avoid bias. However, the system would work for any such objects as long as the object detector can detect them. We experiment with a total of 21 object classes.
\paragraph{Room types}
We experiment with three types of rooms -- kitchen, living room, and bedroom available in ai2thor. We omit bathrooms from our evaluation as the entire room is often visible from a single view. However, our system would work as it is for smaller rooms as well. We sample 5 rooms from the bedrooms as a validation set to optimize the hyper-parameters. For the test data, we initially sample 17 rooms from the kitchen and 17 rooms from living rooms in their default configuration. In the default configuration, very few objects occur in multiple instances, which leads to fewer ambiguities. Thus we spawn copies of different object instances in random but visible locations %\footnote{\url{https://ai2thor.allenai.org/ithor/documentation/objects/domain-randomization/}} 
and further obtain 68 different room configurations, each containing multiple copies ($\leq5$) of a particular object. There are 167 observations on average per room, varying with the room size ($\sigma=108.3$).
\paragraph{Instructions}
We automatically generate the instructions by generating referring expressions and putting them in instruction templates of different verbs. For a given room, for each object present in the room, we construct a referring expression using the simulator-provided metadata. Specifically, for each object class, we construct three types of referring expressions.
\begin{itemize}[leftmargin=*]
\item Referring with a self attribute, e.g., \textit{pick up a plastic cup}.
\item Referring with both self and relational attribute, e.g., \textit{take the plastic cup on the table}.
\item Referring only with the object name, e.g., \textit{bring a cup}.
\end{itemize}

\begin{table}[]
\small
    \caption{F1 scores of label prediction by Bi-LSTM and BERT-based sequence taggers. Boldface numbers are the highest in a row.}
    \centering
    \begin{tabular}{c|c|c}
    \hline
    \textbf{Label} & \textbf{Bi-LSTM tagger} & \textbf{BERT tagger} \\ \hline
    B-$r(g)$    &0.94     &\textbf{0.97}  \\ 
    I-$r(g)$    &\textbf{0.82}       &0.80   \\ 
      B-$av_R$      &0.96       & \textbf{0.97}   \\ 
      I-$av_R$      &0.67       & \textbf{0.84}    \\ 
        B-color $\in at_S$       &\textbf{0.94}        &0.92   \\ 
        I-color $\in at_S$      & 0.84      &\textbf{0.86}       \\ 
          B-material $\in at_S$         &0.94        &\textbf{0.95}    \\ 
            B-is-near $\in at_R$        &0.89         &\textbf{0.93}    \\ 
              B-is-on $\in at_R$        &0.97        & \textbf{0.99}   \\ 
               B-is-at $\in at_R$       &0.89         &\textbf{0.95}    \\ \hline
    Weighted avg.       &0.93       &\textbf{0.94}  \\ \hline
    \end{tabular}
    \label{tab:tagging-result}
%\vspace{-0.6cm}
\end{table}
\subsection{Evaluation of phrase-to-graph network}
We automatically generate pairs of text and labels to train the transformer-based sequence tagger. We have defined templates with several variants of surface forms of instructions and referring expressions with slots for self and relational attributes. Then permutations of the templates are generated by selecting random tokens from sets of pre-defined attributes. We have constructed the text samples with two self attribute types - \textit{color} and \textit{material}; and three relational attribute types - \textit{is-near, is-on} and \textit{is-at}. A total of 6305 text-label sequence pairs are generated. A 20\% of this data is kept as a validation set and the rest is used to fine-tune the BERT transformer along with the tagger network. We train for 15 epochs with the learning rate $5e^{-5}$, as recommended in~\cite{lu2019vilbert}. We test the network on manually annotated test data, which contains 110 pairs of text and label sequences with a total of 590 token-label pairs to predict. We also experiment with a similar Bi-LSTM network with pre-trained GloVe embedding and found the BERT-based approach generally works better. The results are summarized in Table~\ref{tab:tagging-result}. The BERT-based model has 109,492,237 parameters and it takes approximately 32 ms on a 3.70 GHz CPU for a single inference, while the Bi-LSTM model has 131,634 parameters and takes about 3 ms. Thus, the Bi-LSTM model can be used to prefer faster execution over accuracy. We further evaluate the graph construction on the downstream tasks in Section~\ref{sec:e2e-result}.

\subsection{Evaluation of multi-view aggregation algorithm}
In the following, we present the results of object instance detection from the metric grid map. We primarily use the ground truth (GT) 2D object detector in AI2Thor. However, to emulate the behavior of a practical object detector, we model and apply several types of errors to the GT object detection.

\paragraph{Centroid Shift ($E_{CS})$}
Even though the object class is detected correctly, standard object detectors can still fail to recognize the object boundaries perfectly. To emulate this, we randomly shift the centroid of a GT bounding box, keeping its shape intact. We randomly sample a shift in pixels w.r.t. the bounding box's area from a normal distribution ($\mu_C,\sigma_C$). Then we select a random direction from the four quadrants and apply the shift.
\paragraph{Shape Distortion ($E_{SD}$)}
To further analyze the error object boundary prediction, we randomly increase and decrease the size of the GT bounding box, sampling the percentage of change from the normal distribution ($\mu_S,\sigma_S$).
\paragraph{False Negatives ($E_{FN}$)}
We emulate false negatives, i.e., failing to predict an object, by randomly deleting GT bounding boxes with a probability $P_{FN}$.
\paragraph{False Positives ($E_{FP}$)}
Similarly, we emulate false positives, i.e., falsely predicting an object even though it is not present. by randomly introducing bounding boxes with with a probability $P_{FP}$. Firstly, we sample a different room in AI2Thor. Then we randomly select a caption and its bounding box from a random observation in the sampled room and overlay it on the current frame.
\begin{table}[]
\small
    \caption{F1 scores of object instance counting for GT object detection and different error models. The error model parameters are- 
    $(\mu_C=0.2,\sigma_C=0.04),(\mu_S=0.2,\sigma_S=0.04),(P_{FN}=0.15),(P_{FP}=0.15).$
    }
    \centering
    \begin{tabular}{|c|c|c|p{0.7cm}|p{1.6cm}|c|}
    \hline
    \# instances & GT & $ E_{CS} $ & $E_{CS}$, $E_{SD}$ & $E_{CS}$, $E_{SD}$, $E_{FN}$ & $E_{FP}$ \\ \hline
    1  &0.97    &0.95   &0.95   &0.95   &0.88 \\ \hline
    2   &0.88   &0.84   &0.87   &0.81   &0.58 \\ \hline
    3   &0.84   &0.83    &0.80    &0.76   &0.53 \\ \hline
    %4   &0.55   &0.56    &0.58   &0.51   &0.01    \\ \hline
    %5   &0.67   &0.67    &0.67   &0.40   &0.01    \\ \hline
    %AVG.    &0.91   &0.89   &0.90   &0.87   &0.74   \\ \hline
    AVG.    &0.91   &0.89   &0.90   &0.87   &0.74   \\ \hline
    \end{tabular}
    \label{tab:counting-result}
%\vspace{-0.2cm}
\end{table}

\begin{figure}[t!]
    \centering
    \includegraphics[width=1.0\linewidth]{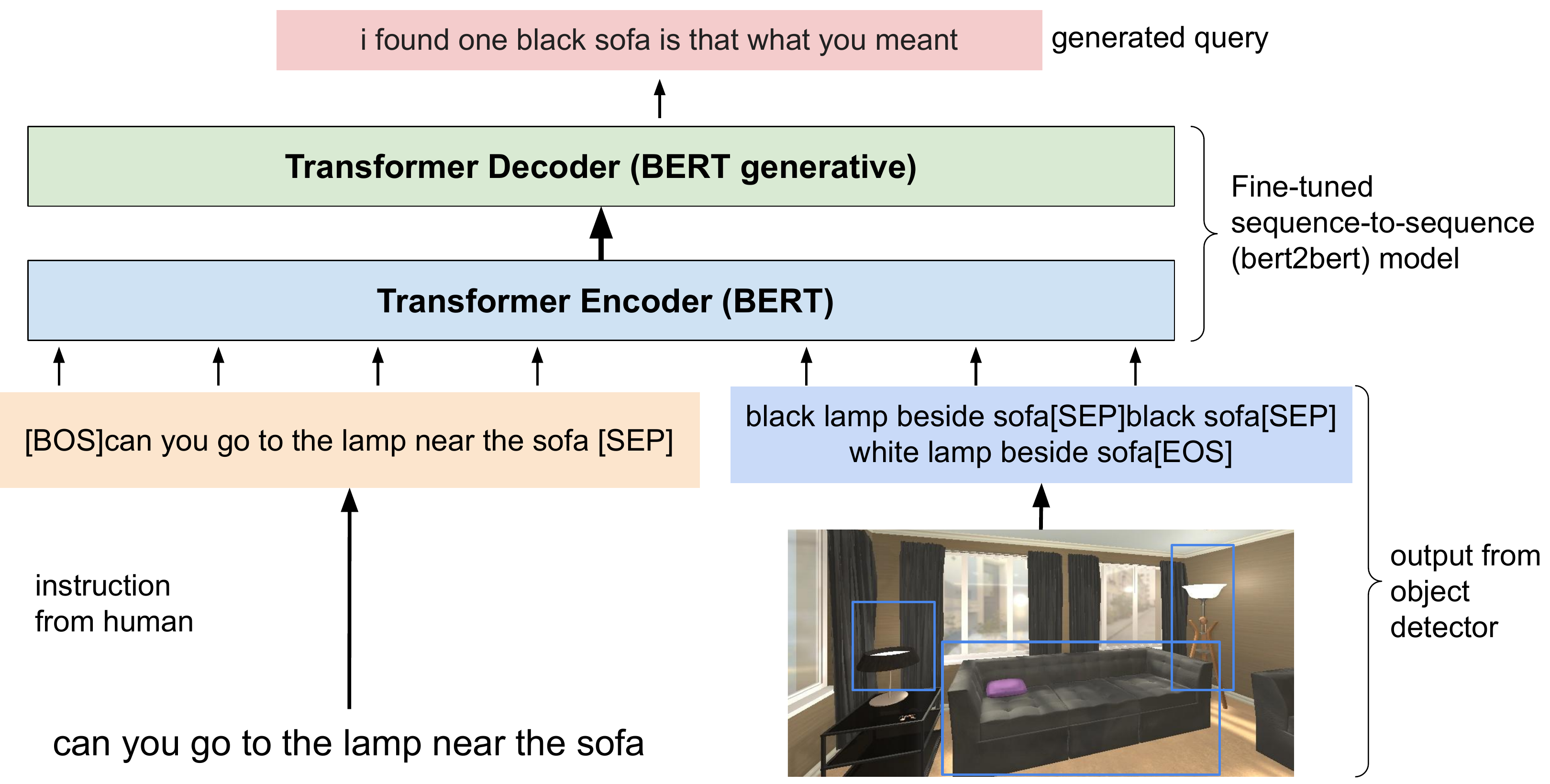}
    \caption{A baseline system for ambiguity detection and query generation is created by fine-tuning a bert-2-bert based encoder-decoder model.}
    \label{fig:visitron_model}
%\vspace{-0.6cm}
\end{figure}

The goal of multi-view aggregation is to correctly find the unique object instances. If the algorithm finds more object instances than the ground truth, it can lead to ambiguity. Similarly, if the algorithm suggests less number of object instances than the actual, it may not capture the ambiguity. Table~\ref{tab:counting-result} shows the F1 scores for object instance counting after aggregation for ground truth object detector and various noisy detectors that we described earlier. It is noted that even if the GT object detector is used, the number of object count is impacted due to the noise introduced by the aggregation mechanism. However, the noisy object detector has minimal impact as the aggregation almost cancels out the noise. Even though the false positive (FP) scenario is the worst affected, state-of-the-art object detectors have minimal FP cases.

%\vspace{-0.15cm}
\subsection{Evaluation of the overall system}
\label{sec:e2e-result}
We compare our approach against a baseline that can detect ambiguity and generate queries only from a single observation. Following the similar network architecture proposed by Shrivastava \textit{et al.}~\cite{shrivastava2021visitron}, we fine-tuned an end-to-end sequence-to-sequence network as shown in Fig.~\ref{fig:visitron_model}. We used a pre-trained BERT model to fine-tune the baseline system. The training data source is the same as our model. We compared the baseline system with \system in terms of ambiguity and generated query. The generated queries are compared against a reference query. Firstly, we calculate the BLEU score to find the token level similarity of the generated query. Then, we find semantic similarity (BERT score) by using a BERT-based sentence level similarity score. Then, we calculate the F1 score for finding the ambiguity (ambiguity accuracy -- AA). However, an ambiguity may be erroneously triggered for a completely different reason. This may lead to incorrect query generation and failure in disambiguation. Thus, we manually checked the accuracy of the generated query (QA) as well. We used the GT object detector in combination with $E_{CS}+E_{SD}+E_{FN}$ errors. A total of 600 image sequence and instruction pairs are used for evaluation.

\begin{table}[t!]
\small
    \caption{For a single image, comparison of \system with respect to the baseline system in terms of describing the ambiguity.}
    \centering
    \begin{tabular}{|p{1.1cm}|p{0.8cm}|p{0.8cm}|p{1.7cm}|p{1.7cm}|}
    \hline
    %metric & BLEU score & BERT score & AA & QA \\ \hline
    metric & QA & AA & BLEU score & BERT score  \\ \hline
    Baseline & 0.45 & 0.92  & 0.75  & 0.82 \\ \hline
    %Baseline & 0.75  & 0.82   & 0.92   & 0.45 \\ \hline
    %\system & \textbf{0.81}   & \textbf{0.90}   & \textbf{0.94} & \textbf{0.85} \\ \hline
    \system  & \textbf{0.85} & \textbf{0.94} & \textbf{0.81}   & \textbf{0.90}  \\ \hline
    \end{tabular}
    \label{tab:system_sccuracy_single_image}
    %\vspace{-0.2cm}
\end{table}
\begin{table}[t!]
\small
    \caption{Performance of \system (end-to-end) with aggregated data from multiple views as compared to a single-view image with maximum number of target object instance(s).}
    \centering
    \begin{tabular}{|p{3.5cm}|p{0.6cm}|p{0.6cm}|p{0.8cm}|p{0.8cm}|}
    \hline
    %metric & BLEU score & BERT score & AA & QA \\ \hline
    metric  & QA & AA & BLEU score & BERT score \\ \hline
    %Baseline (single view) & 0.64 & 0.70   & 0.90   & 0.37\\ \hline
    Baseline (single view)  & 0.37  & 0.90 & 0.64 & 0.70  \\ \hline
    %\system (single view) & 0.71  & 0.84  & 0.87 & 0.69 \\ \hline
    \system (single view) & 0.69  & 0.87 & 0.71  & 0.84\\ \hline
    %Baseline (multi-view caption aggregated) & 0.26  & 0.36  & 0.51 & 0.07 \\ \hline
    Baseline (multi-view caption aggregated) & 0.07 & 0.51 & 0.26  & 0.36 \\ \hline
    %Baseline (multi-view our aggregator) & 0.56  & 0.69  & 0.82 & 0.67 \\ \hline
    Baseline (multi-view our aggregator) & 0.67 & 0.82 & 0.56  & 0.69 \\ \hline
    %\system (multi-view) & \textbf{0.77}  & \textbf{0.87}  & \textbf{0.91} & \textbf{0.77} \\ \hline
    \system (multi-view)  & \textbf{0.77} & \textbf{0.91} & \textbf{0.77}  & \textbf{0.87} \\ \hline
    \end{tabular}
    \label{tab:system_sccuracy}
    %\vspace{-0.6cm}
\end{table}
Now, imagine a setup where the robot is either static or a single view is sufficient for grounding the referred object. In such a scenario, how good are our ambiguity detection and the generated query? Table~\ref{tab:system_sccuracy_single_image} summarizes the results. Notice that even if the baseline system detects ambiguity with high precision, it fails to generate an accurate query for the same. The reason is, the query generation draws the tokens from the language model in the decoder. Hence, it prefers most likely token rather than the context-specific token. So, even if the model predicts a grammatically and semantically accurate query, the information conveyed is wrong. This poor accuracy would lead to poor disambiguation. 

In Table~\ref{tab:system_sccuracy}, we summarize the results for a more practical scenario where multiple views are required to assess the ambiguity. We evaluate the baseline system in three configurations- a) single view captions, b) multi-view captions naively concatenated with a delimiter, and c) output of our multi-view aggregation converted to natural language text. The single view is chosen randomly from the observations having maximum referred object instances. In single view, both the baseline and our system perform poorly in the accuracy of the query, whereas our system performs way better than the baseline. However, after aggregation, the accuracy of our system improves significantly. Using our aggregation strategy also improves the baseline's performance, whereas the naive concatenation does not give meaningful  queries.

%\vspace{-0.15cm}
\subsection{Discussion}
Even though our system shows a very promising result in embodied object disambiguation, several future extensions are possible. Firstly, the system should be extended to close the loop of dialogue, i.e., after raising the query, the system should accept the user input and resolve the ambiguity. Secondly, the quality of the generated query as well as the perception of accuracy can be assessed and rectified by conducting user participation and survey. Third, the accuracy of unique object detection depends on the area exploration strategy. We have used a pseudo-random exploration strategy to exhaustively cover the area. In the future, we shall work on an exploration strategy to cover the area with minimal but sufficient views. Fourth, the system can be extended by integrating a practical caption generator. Fifth, a pre-trained phrase-to-graph network can be fine-tuned using human annotations for domain adaptation. The tag-to-graph grammar can also be extended to take input from a dependency parser. Lastly, our current aggregation algorithm is not evaluated against stacked objects, which can be done easily.

%% file: 6_conclusions.tex
\section{CONCLUSIONS}
In this work, we present a novel and effective system, \system, to tackle the object disambiguation task for an embodied agent. With the phrase-to-graph network, we can convert any natural language object description into a semantic representation (object graph). This not only provides a formal representation of the referred object and object instances but also helps to find the ambiguity using discrimination. We propose a real-time multi-view aggregation algorithm that processes multiple observations from the environment and finds the unique instances of the referred object. We have also conducted extensive experiments to study \system and compared its efficacy with a baseline system. The main motivation behind this work is not just finding the ambiguity, but qualifying it with accurate, context-specific information so that it is sufficient for a human being to come up with a reply towards disambiguation.